%% file: bmvc_review.tex
\documentclass{bmvc2k}

\usepackage{graphicx}
\usepackage{amsmath}
\usepackage{bm}
\usepackage{amssymb}
\usepackage{enumitem}
\usepackage{makecell}
\usepackage{tabularx}
\usepackage{booktabs}
\usepackage{pifont}
\usepackage{wrapfig}
\usepackage{float}

\usepackage{array}
\newcommand{\PreserveBackslash}[1]{\let\temp=\\#1\let\\=\temp}
\newcolumntype{C}[1]{>{\PreserveBackslash\centering}p{#1}}
\newcolumntype{R}[1]{>{\PreserveBackslash\raggedleft}p{#1}}
\newcolumntype{L}[1]{>{\PreserveBackslash\raggedright}p{#1}}
\usepackage{booktabs}

\usepackage[font={color=bmv@captioncolor}]{caption}
\usepackage[capitalize]{cleveref}
\crefname{section}{Sec.}{Secs.}
\Crefname{section}{Section}{Sections}
\Crefname{table}{Table}{Tables}
\crefname{table}{Tab.}{Tabs.}

\newcommand{\fakeparagraph}[1]{\noindent\textbf{#1}}
\title{Privacy Vulnerability of Split Computing to Data-Free Model Inversion Attacks}

\addauthor{Xin Dong}{xindong@g.harvard.edu}{1}
\addauthor{Hongxu Yin}{dannyy@nvidia.com}{2}
\addauthor{Jose M. Alvarez}{josea@nvidia.com}{2}
\addauthor{Jan Kautz}{jkautz@nvidia.com}{2}
\addauthor{Pavlo Molchanov}{pmolchanov@nvidia.com}{2}
\addauthor{H.T. Kung}{kung@harvard.edu}{1}

\addinstitution{
 Harvard University\\
 John A. Paulson School of Engineering and Applied Sciences\\
 Allston, USA
}
\addinstitution{
 NVIDIA\\
 2788 San Tomas Expy \\ 
 Santa Clara, USA
}

\runninghead{Dong, Yin, and et al.}{Data-Free Model Inversion to Split Computing}

\def\eg{\emph{e.g}\bmvaOneDot}

\def\ie{\emph{i.e}\bmvaOneDot}

\def\etal{\emph{et al}\bmvaOneDot}
\newcommand{\method}{DCI\xspace}

\begin{document}

\maketitle
% \vspace{-4mm}
\begin{abstract}
Mobile edge devices see increased demands in deep neural networks (DNNs) inference while suffering from stringent constraints in computing resources.
Split computing (SC) emerges as a popular approach to the issue by executing only initial layers on devices and offloading the remaining to the cloud. Prior works usually assume that SC offers privacy benefits as only intermediate features, instead of private data, are shared from devices to the cloud.
In this work, we debunk this SC-induced privacy protection by (i) presenting a novel data-free model inversion method and (ii) demonstrating sample inversion where private data from devices can still be leaked with high fidelity from the shared feature even after tens of neural network layers. We propose Divide-and-Conquer Inversion (DCI) which partitions the given deep network into multiple shallow blocks and inverts each block with an inversion method. Additionally, cycle-consistency technique is introduced by re-directing the inverted results back to the model under attack in order to better supervise the training of the inversion modules. In contrast to prior art based on generative priors and computation-intensive optimization in deriving inverted samples, DCI removes the need for real device data and generative priors, and  completes inversion with a single quick forward pass over inversion modules. For the first time, we scale data-free and sample-specific inversion to deep architectures and large datasets for both discriminative and generative networks. We perform model inversion attack to ResNet and RepVGG models on ImageNet and SNGAN on CelebA and recover the original input from intermediate features more than 40 layers deep into the network. Our method reveals a surprising privacy vulnerability of modern DNNs to model inversion attacks, and provides a tool for empirically measuring the amount of potential data leakage and assessing the privacy vulnerability of DNNs under split computing.

\end{abstract}

\vspace{-4mm}
\section{Introduction}
\vspace{-1mm}
In a modern Internet of Things system, distributed intelligent devices and a server in the cloud form a hardware backbone to realize complex deep learning systems~\cite{teerapittayanon2016branchynet,teerapittayanon2017distributed,verbraeken2020survey}. With such a system, it is natural to split a deep learning workload between the devices and the server in support of distributed computing 
which is known as as \textit{split computing} (SC)~\cite{matsubara2020head}.
Specifically, for a DNN with $M$ layers, devices execute the first $L$ layers and send the resulting features to the server, which then completes the remaining $(M-L)$ layers. In prior literature~\cite{teerapittayanon2016branchynet,matsubara2021split,thapa2022splitfed}, it is assumed that SC protects data privacy given the difficulty of recovering original data from features when $L$ is large. In this work, however, we demonstrate that it is still possible to invert a sub-network of many layers 
without depending on any real training data~\citep{RN58,zhang2020secret} or input-space priors (\eg, pre-trained generative models~\cite{zhang2020secret,struppek2022plug}). 

In numerous real-world SC applications~\citep{kang2017,yao2020deep,dong2022splitnets}, an off-the-shelf pre-trained DNN is split into two parts and then deployed to devices and the cloud separately. Common threat 

\input{figs/sc/sc_fig}
\noindent models for recovering raw data from a device (\cref{fig:teaser}) therefore assume that the adversary has access to the on-device sub-network (termed as target model) as well as the shared feature activations. The adversary can be considered as either an honest-but-curious server or an eavesdropper over the communication channel. 
Prior attack methods~\cite{RN58,yang2019neural}, in addition, assume that the dataset used for training the target model (or a similar dataset~\cite{zhang2020secret}) are available for the adversary to train its inversion model. In practice, however, it could be unlikely for an adversary to know or obtain these real data. Moreover, a line of approaches~\cite{zhang2020secret,li2022auditing,struppek2022plug} rely on pre-trained generative models~\cite{brock2018large,karras2020analyzing} to perform inversion. Due to a similar reason, it may be challenging for the adversary to obtain such generative models without having information on the real data. 

In this work, we focus on a white-box and data-free model inversion attack. Previous theoretical~\cite{Arora2015,Gilbert2017,QiLei2019} and empirical~\cite{RN58,teterwak2021understanding,hatamizadeh_2022} studies have shown that inversion attack becomes more difficult as the depth of the network that computes the feature increases. To overcome the increased complexity and non-linearity of the target model as the depth increases, we propose a \textit{divide-and-conquer inversion} method that partitions the inversion into sequential layer/block-wise inversion sub-problems. For each sub-problem, an inversion module is built and optimized to reverse the input-output function of the corresponding target layer/block. 
To support optimization of inversion modules, we exploit synthetic data generated via minimizing the discrepancy between (randomly initialized) dummy input's feature statistics and the corresponding statistics stored in batch normalization (BN) layer~\citep{santos2019learning,yin2020dreaming,xu2020generative}. 
With synthetic data proxies, we are able to optimize the inversion model to enforce feature embedding similarity with respect to the original counterpart. We call it cycle \textit{consistency-guided inversion}. Combining aforementioned techniques enables us to successfully scale zero-shot model inversion to deep architectures and large datasets without strong assumptions on model or dataset priors.

In summary, we make the following contributions: (1) We propose an inversion method based on a divide-and-conquer inversion (DCI) and a cycle-consistency loss on synthetic data. (2) We demonstrate that the method generalizes to both discriminative and generating models, even for large datasets~(\textit{i.e.}, ImageNet~\citep{deng2009imagenet} and CelebA~\citep{liu2015faceattributes}), surpassing state-of-the-art baselines. (3) We demonstrate that \method can be used as a tool to analyze the inversion difficulty at different layer depths as well as the trade-off between utility and defense.

% \vspace{-3mm}
%-------------------------------------------------------------------------
\section{Related Work}

% \subsection{Model Inversion Attacks}
Prior model inversion attacks have been mainly pursued in
three directions: analytical inversion, generative inversion, and input optimization-based inversion. This work aims to relax some strong assumptions made by prior arts, such as the requirement for real data, pretrained image priors, and intensive optimization. These assumptions may cause attacks to be time-  and resource-consuming, inflexible and susceptible to distribution shifts between data used by the target model and data used for training the inversion model. 

\input{tabs/related_work}

\noindent\textbf{Analytical Inversion.}
Analytical inversion studies the theoretical invertibility of neural networks~\cite{arora2014provable}. \cite{Arora2015,Gilbert2017} theoretically show that the approximate reverse of a layer by transposing its weights matrix, based on the assumption that its weights are random-like, which is not always true for a pre-trained network~\cite{Cai2019WeightNB,Li2020AdditivePQ}.  
Lei~\etal~\cite{QiLei2019} indicates that a layer can be inverted via linear programming assuming its output dimension is greater than twice its input dimension. However, this assumption does not hold for most classification models~\cite{He2015}.
Despite theoretical soundness of analytical inversion, there is still a discrepancy between the setups of theories and those of contemporary pre-trained models. Moreover, these methods cannot scale to large network sizes beyond MNIST~\cite{lecun1998mnist}/CIFAR-10~\cite{krizhevsky2009learning} level, and the recovered images are noisy~\citep{Arora2015}.

\fakeparagraph{Generative Inversion.} 
Generative inversion aims to learn (or leverage) a generative model that reverses the input-output mapping of the target model. 
`Plug\&Play' methods~\cite{Nguyenplugplay2017,teterwak2021understanding} utilize a pre-trained generator (\textit{e.g.}, BigGAN~\cite{brock2018large}) as the `image prior' to generate images that maximize the activation of the target model. If pre-trained generators are unavailable, a generative model will be trained on the original dataset of the target model~\cite{RN58,zhang2020secret}. 
The dependence on the pre-trained generator, the original dataset, and challenging adversarial learning~\cite{arjovsky2017wasserstein,kodali2017convergence} limits the practicability of generative inversion. In addition, these methods focus on class-specific instead of accurate sample-specific inversion. For instance, given the feature of a husky dog image, generative inversion may recover a Teddy dog image because the generator adds some semantic-related (\eg, a dog in general) but sample-unrelated (\ie, the specific dog) information during inversion~\cite{zhang2020secret,struppek2022plug}. This problem would be exacerbated if the data used to train the generator had a different distribution than the original data.

\fakeparagraph{Input Optimization Based Inversion.}
\cite{cai2020zeroq,haroush2020knowledge,chawla2021data} show that one can reveal certain information of training set by optimizing a dummy input to match pre-stored feature statistics~\citep{yin2020dreaming} or maximum activation of neurons~\cite{mordvintsev2015deepdream}. However, these methods are all designed to leak arbitrary training samples instead of targeted sample inversion. In addition, these methods remain computationally heavy.

\fakeparagraph{Gradient Inversion.} Training neural networks requires gradient estimation from a batch of data. \cite{yin2021see,zhu2019deep,li2022auditing,geiping2020inverting,hatamizadeh2022gradvit} demonstrate that sensitive input data can be leaked given gradients. Different from gradient inversion which mainly targets on recovering input from gradients during training, we focuses on reconstructing input from feature map during inference (especially for spilt computing scenarios). More detailed comparison is provide in~\cref{tab:key_difference}.

% \vspace{-1mm}
\section{Method}
\label{sec:method}
Before describing the proposed method, we introduce some notations. Consider a multi-layer on-device (discriminative or generative) sub-network with $L$ layers, $\mathcal{F}_{1:L}(\mathbf{x}) := \mathcal{F}_L \circ  \mathcal{F}_{L-1} \circ \dots \circ \mathcal{F}_1(\mathbf{x})$. 
$\mathcal{F}_k$ represents the $k$-th parameterized layer which may include its associated batch normalization~\cite{ioffe2015batch} and activation.
$\mathcal{F}_{k:l}$ is the sub-network containing $k$-th to $l$-th layers inclusively. 
In this work, we consider the following question on model inversion attacks~\cite{fredrikson2015model,wang2021variational}: given a sub-network $\mathcal{F}_{1:L}$ pre-trained on dataset $\mathcal{D}$, is it possible to learn its approximate reverse counterpart $\mathcal{F}_{1:L}^{-1}$ (where $\mathcal{F}_{1:L}^{-1}: \mathcal{F}_{1:L}^{-1}(\mathcal{F}_{1:L}(\mathbf{x}))=\mathbf{x}'\approx \mathbf{x}$) without access to $\mathcal{D}$? 
\subsection{Divide-and-Conquer Inversion}
\label{sec:dci}
DNN consists of blocks of layers, with each layer producing an intermediate feature map that is consumed by the next layer as input.
Prior research has demonstrated that neural networks produce more abstract feature maps as the depth increases~\cite{bianchini2014complexity,eldan2016power,telgarsky2016benefits,raghu2017expressive}. Consequently, existing studies have reported the difficulty of inverting deep models~\cite{Gilbert2017,RN58,Dosovitskiy2016,Fangchang2018}.
To circumvent the difficulty of approximating all stacked layers, we first split the overall inversion problem into several layer-(or block-)wise inversion sub-problems before combining them. To this end, we present a simple yet effective inversion strategy called \textbf{D}ivide-and-\textbf{C}onquer \textbf{I}nversion~(DCI) that progressively inverts the computational flow of DNNs. Comparing to end-to-end inversions~\cite{zhang2020secret,wang2021variational}, DCI has two advantages: 
(1) a single layer~(or block) has less non-linearity and complexity, and is therefore easier to invert; (2) DCI provides richer supervision signals across layers~(or blocks), whereas the overall inversion only utilizes supervision at the two ends (\ie, the input and output) of the target model. 

Starting from the first layer, DCI inverts each layer/block $\mathcal{F}_k$ with two goals: (i) inverting current target layer $\mathcal{F}_k$ with an inversion module $\mathcal{F}^{-1}_k$ of similar size, and (ii) fine-tuning the newly inverted module $\mathcal{F}_k^{-1}$ to ensure that it works well jointly with all previously inverted layers $\mathcal{F}_{1:(k-1)}^{-1}$. 
The necessity of the second objective comes from the non-exact inversion of each layer/block. Otherwise, a small error could be amplified along layers~\cite{dong2017learning}. 
Therefore we refine inverted modules to minimize the accumulated error. 
According to existing literature~\cite{zhu2017unpaired,Richard2018} We use $\ell_1$ distance to measure the reconstruction loss.
For the first objective, we minimize the layer reconstruction loss,
\vspace{-1mm}
\begin{align}
    \mathcal{L}_\text{layer} = \left\|\mathbf{u}_k - \mathcal{F}_k^{-1}\left(\mathcal{F}_k\left(\mathbf{u}_k\right)\right)\right\|_1,\quad \mathbf{u}_k = \mathcal{F}_{1:(k-1)}(\mathbf{x}),\vspace{-1mm}
    \label{equ:layer_reconst}
\end{align}
where $\mathbf{u}_k$ is the input of the target layer/block. As for the second objective, we aim to ensure the reconstruction quality of inversion modules up to the layer $k$ such that \scalebox{0.8}{$\mathcal{F}_{1:k}^{-1}=\mathcal{F}_k^{-1}\circ \mathcal{F}_{1:(k-1)}^{-1}$}. This translates into
% measuring how close the inverted input 
minimizing the distance between inverted input \scalebox{0.8}{$\mathbf{x}' = \mathcal{F}_{1:k}^{-1}\left(\mathcal{F}_{1:k}\left(\mathbf{x}\right)\right)$} 
to the original input $\mathbf{x}$, where we introduce another loss term for this purpose:
\vspace{-2mm}
\begin{equation}
    \mathcal{L}_\text{img}(\mathbf{x}, \mathbf{x}')=\|\mathbf{x}-\mathbf{x}'\|_1,\quad \mathbf{x}' = \mathcal{F}_{1:k}^{-1}\left(\mathcal{F}_{1:k}\left(\mathbf{x}\right)\right).
\label{equ:img_l1}
\end{equation}

\noindent\textbf{Cycle-consistency Guided Inversion.}
\label{sec:cycle}
We further re-exploit the target model for stronger inversion guidance amid the unique setup of the inversion problem. As we reverse the computation of the target model, this forms a natural loop with the original computation flow. 
Inspired by perceptual metric~\cite{Richard2018} and cycle-consistent image translation~\cite{zhu2017unpaired}, we measure the quality of reconstructed inputs by re-checking them with the target model. 

The main idea is that if the reconstructed input is faithfully and semantically close to the original input then the target model should produce similar, if not exact, feature responses at all layers. To this end, we feed the reconstructed input back to the direct (target) model, and minimize the distance between features of the reconstructed input and original input at various depths. 
We refer to this as cycle consistency that is defined as follow: 
\vspace{-1mm}
\begin{equation}
    \displaystyle{
    \mathcal{L}_\text{cyc}(\mathbf{x}, \mathbf{x}') = \Sigma_{l=1}^L \|\mathcal{F}_{1:l}(\mathbf{x})-\mathcal{F}_{1:l}(\mathbf{x}')\|_1}.
    \label{equ:cyc}
    \vspace{-1mm}
\end{equation}
This enables a better utilization of the features from the original input twice to provide richer supervision: (\textbf{i}) In \cref{equ:layer_reconst}, we use features of the original input $\mathcal{F}_{1:(k-1)}(\mathbf{x})$ as the reconstruction objective; (\textbf{ii}) The cycle consistency loss (\cref{equ:cyc}) also uses features of the original input as a reference and enforces the inverted input similar to those of the original input. % (i.e., \ding{183} in~\cref{fig:cycle}). 

With all the above losses, the overall optimization objective for an inversion layer can thus be expressed as $\mathcal{L}_k=\mathcal{L}_\text{layer}+\mathcal{L}_\text{img}+\mathcal{L}_\text{cyc}$.

\subsection{Data Sampling}
\label{sec:data_sampling}
One remaining challenge to optimize inversion modules is how to obtain input data $\mathbf{x}$. When the target model is a generative model (\textit{i.e.}, the generator of a generative adversarial network~\cite{aggarwal2021generative}), obtaining input data is as simple as randomly sampling latent codes from a Normal distribution~\cite{arjovsky2017wasserstein}. However, when the target model is a discriminative model, it is infeasible to sample data from the original data distribution under the data-free setting.

Inspired by adversarial-free generative models~\cite{li2017mmd,binkowski2018demystifying} and data-free knowledge distillation~\cite{yin2020dreaming,xu2020generative}, we leverage these techniques to generate a small dataset for training of inversion modules. 
% The method we choose 
In particular, we generate a batch of data by minimizing the difference between features statistics of the synthetic data $\hat{\mathbf{x}}$ and statistics stored in batch normalization (BN): 
\vspace{-2mm}
\begin{equation}
{\text{min}}_{\hat{\mathbf{x}}}{\ \sum\nolimits_{l}|| \mu_{l}(\mathbf{\hat{x}}) - \text{BN}_{l}( \mu)||_2 + \sum\nolimits_{l}|| {\sigma^2_l}(\mathbf{\hat{x}}) - \text{BN}_{l}(\sigma^2) ||_2}+\lambda\mathcal{R}_\text{img}(\hat{\mathbf{x}}).\vspace{-1mm}
\end{equation}
Here, $\mu_l(\hat{\mathbf{x}})$ and $\sigma^2_l(\hat{\mathbf{x}})$ are features statistics computed at the $l$-th layer. $\text{BN}_{l}( \mu)$ and $\text{BN}_{l}(\sigma^2)$ are their corresponding statistics stored in a BN layer.  $\mathcal{R}_\text{img}$ is the image regularization like total variance~\cite{guo2017countering} to make derived images more natural~\cite{yin2020dreaming}.
Minimizing the difference of features' statistics 
is equivalent to reducing the probability distance~\cite{li2017mmd} between distributions of synthetic data and real data~\cite{choi2020data}.
Thus, the synthetic data can be treated as data from a distribution that is similar to the underlying distribution of real data and thus can serve as a reasonable proxy for training of our inversion modules. 

Note that there is still a gap between the synthetic and original real data. Different from our approach, prior generative inversions cannot utilize this synthetic data because they are sensitive to this gap due to the following reasons:
(1) Generative inversions~\cite{RN58,teterwak2021understanding} train a generator (in a GAN framework) as the inversion model, whereas the GAN training is very sensitive and challenging given only synthetic data~\cite{kodali2017convergence,tran2021data,zhao2020diffaugment}. The resulting efficacy falls short to DCI, as we will show later in~\cref{sec:exp_comp}.
% \vspace{-1mm}
(2) Prior works~\cite{Nguyenplugplay2017,RN58,teterwak2021understanding} require the whole training set~(\textit{e.g.}, more than $1$M images) to train the GAN. Yet deriving massive synthetic data remains slow~\cite{yin2020dreaming}. 
% \end{itemize}
In contrast, without any adversarial learning, DCI does not have the above disadvantages when using synthetic data because of its progressive inversion nature and cycle consistency supervision.
DCI is data-efficient and can be enabled by only $10$K synthetic images. In contrast, generative methods requires the access to the original ImageNet training dataset containing 1.28M real images~\cite{zhang2020secret,Nguyenplugplay2017}. 

\section{Evaluation}
In this section, we demonstrate the efficacy of our inversion attack to discriminative and generative models. Without any strong assumptions such as real data, costly input optimization, or pre-trained image priors, we significantly scale model inversion attack to very deep architectures (\textit{e.g.}, RepVGG-A0, ResNet-18, ResNet-50, and SNGAN), high resolution input (\textit{e.g.}, $224\times224$px for ImageNet and $128\times128$px for CelebA), and different pre-training methods for the target model (\textit{e.g.}, supervised and self-supervised). Due to page limit, we present more implementation details (\cref{sec:implement_details}), additional inversion results (\cref{sec:addition_inversion}), and ablation studies (\cref{sec:ablation}) in the appendix.  

\input{figs/moco_inv/only_resnet50}
\subsection{Inversion Attack to Deep Models on the ImageNet Dataset}
\label{sec:exp_img}
\noindent\textbf{Inversion Attack to RepVGG-A0.}
We consider the inversion of RepVGG~\cite{ding2021repvgg}, which is one of the state-of-the-art classification models with a deep architecture. It has a ResNet-like~\cite{He2015} multi-branch topology during training, and a mathematically equivalent VGG-like~\cite{simonyan2014very} inference-time architecture, which is achieved by layer folding. Specifically, we consider RepVGG-A0~\cite{ding2021repvgg} with $22$ convolution blocks, each of which is comprised of a convolution, ReLU, and BN layer, yielding
a $72.41\%$ top-1 accuracy on ImageNet~\cite{ding2021repvgg}. 
Using our DCI strategy, We invert one block with an inversion module at a time. Each inversion module has a similar architecture as its target counterpart but with reversed input-output dimensions, and is optimized via Adam~\cite{kingma2014adam} for $6$K iterations. We use $10$K synthetic images
as detailed in~\cref{sec:data_sampling} to train inversion modules.
Inversion attack results are displayed in \cref{fig:repvgg_main}.
Remarkably, our inversion recovers very close pixel-wise proxy to the original image from the $21$-th layer's activation (with spatial size $14\times14$), preserving original semantic and visual attributes, such as color, orientation, outline, and position.

\input{figs/repvgg_main/repvgg_main}

\noindent\textbf{Inversion Attack at Varying Depths.}
We visualize the inverted images from different layer depths of the target model in~\cref{fig:diff_layer_repvgg}. As the depth increases, inverted images become more blurry, losing mo-

\input{figs/layerwise_repvgg/layerwise_repvgg_fig}
\noindent re details. It is caused by information filtering due to the classification nature of the model~\cite{NIPS2014_375c7134}. To quantify the changes, we also provide peak signal-to-noise ratio~(PSNR)~\cite{psnr} and perceptual metric 
%image patch similarity metric
(LPIPS)~\cite{Richard2018} between real and inverted images at different depths. The quantitative results are consistent with the qualitative observations.
An intriguing observation from~\cref{fig:diff_layer_repvgg} is the consistent preservation of high level information across the majority of layers, which challenges prior security arguments in split computing~\cite{kang2017,eshratifar2019bottlenet,dong2022splitnets}. 
As shown in~\cref{fig:diff_layer_repvgg}, we can achieve nearly perfect recovery from outputs of the $5$-th blocks (with spatial size $28\times28$) that already past three stride-$2$ convolutions.
which were regarded as lossy operations.
We find that increasing the depth of on-device sub-network is not always effective in making inversion more difficult. For example, in~\cref{fig:diff_layer_repvgg} (right), features from the $8$-th to $16$-th blocks results in inverted images with comparable quality in terms of PSNR and LPIPS. In addition, we notice an obvious drop of inversion quality (\ie, PSNR) after the second convolution layer which has 48 in and out channels with a stride of 2. This layer repeatedly reduces the feature size from $48\cdot112^2$ to $48\cdot56^2$, resulting in information loss~\cite{ma2019quantifying,zhu2020r} and increased inversion difficulty. The following (\ie, the third) layer has a stride of 1 and the same input and output sizes. According to~\cref{fig:diff_layer_repvgg} (right), this layer may enhance some information~\cite{hermann2020shapes}, resulting in an improved inversion quality.  

In addition, we find that the final two layers dramatically degrade the inversion quality. One can still recognize the class of inverted images after the penultimate (\textit{i.e.}, the $22$-nd convolution) block. However, if we invert features after the final~(\textit{i.e.}, the fully-connected) layer, only the predominant color is discernible as shown in \cref{sec:addition_inversion}. This suggests that class-invariant information is rapidly filtered out towards the end of the model. This is consistent with observations in transfer learning~\cite{NIPS2014_375c7134,li2020rifle,zoph2020rethinking} and self-supervised learning~\cite{he2020momentum,grill2020bootstrap,chen2020big}. 

These results provides several insights to the privacy vulnerability of split computing. For an extremely compact edge device, only few (\textit{e.g.}, 5) blocks can be on the device and the other layers have to be on the cloud. In this case, inverting from the output of the 5-th blocks results in recovered images with many details, \textit{e.g.}, in~\cref{fig:diff_layer_repvgg}, even tiny bottoms and knobs on the oscilloscope are well recognizable. If the edge device has mediate computation/memory capacity, more (\textit{e.g.}, 15) layers can be on the device. Inverting from 15-th block’s output will mainly reveal “large structures” (e.g., outlines, colors, and shapes), also leading to privacy concerns. Both cases debunk privacy of split computing that is presumed safe.

\noindent\textbf{Inversion Attack to ResNets.}
To demonstrate the general applicability of the proposed method, we present inversion results for additional networks on the ImageNet dataset,
encompassing a variety of architectures (ResNet-$18$ and ResNet-$50$) and pre-training methods (standard~\cite{He2015} and self-supervised~\cite{chen2020mocov2}).
Each \texttt{BasicBlock} unit in ResNet-18 comprises 

\input{figs/resnet18_inv/resnet18_inv}
\noindent two convolution layers (with BN and ReLU) and a shortcut connection, and is inverted individually. 
\cref{fig:resnet18_main} depicts the inversion results of pre-trained ResNet-$18$ on ImageNet. With the proposed method, we can recover recognizable input images after up to $7$ blocks
that contain a total of $15$ convolution layers and one maxpooling layer. We observe inversion can still restore images with high fidelity that contain similar visual details as the original images.

We next move to invert pre-trained ResNet-$50$ and investigate both a standard supervised model~\cite{He2015}, and the recent self-supervised model MoCoV2~\cite{chen2020mocov2} in \cref{fig:moco_main}. We observe that a stronger feature extractor (\textit{i.e.}, MoCoV2) preserves more information, resulting stronger inversion, when compared to a standard supervised training of the same network architecture (the second and third rows in \cref{fig:moco_main}). This is consistent with the recent findings in~\cite{yin2021see}, showing that self-supervised models are more vulnerable to gradient inversion attack. These results demonstrate that we are able to recover recognizable input images for up to $42$ convolution layers and one max pooling layer in total.

\subsection{Comparison to Prior Art on Inversion Attack}
\label{sec:exp_comp}
We compare \method to prior model inversion attacks under the data-free setting in~\cref{fig:sota_comp} and~\cref{tab:comp_number}. Among the baselines considered, DDream~\cite{mordvintsev2015deepdream} and DI~\cite{yin2020dreaming} are data-free approaches. They randomly initialize a dummy input and optimize it to recover data samples. Since these methods start from random initialization, it is unlikely to recover a specific instance of input image. Moreover, most of inverted images are visually bizarre. \method utilizes the synthetic data generated by DI~\cite{yin2020dreaming} to optimize inversion modules, and achieves sample-level inversion with significantly improved quality of inverted images. In addition, both DDream and DI require 20K forward and backward passes to optimize inputs – while our method only needs one forward pass through inversion modules, hence is more efficient.
PSiM~\cite{RN58} and GMI~\cite{zhang2020secret} require the original training data of the target model and auxiliary data~\cite{zhang2020secret} to train the inversion model. For a fair comparison, their inversion models are instead trained using synthetic data. As we discussed in~\cref{sec:data_sampling}, these generative inversions do not work well under data-free setting due to their sensitivity to shifts between real and synthetic data distributions.

\subsection{Inversion Attack to GAN on the CelebA Dataset}
\label{sec:exp_gan}

We next demonstrate that our \method can also be applied to generative models. This experiment focuses on the popular SNGAN architecture that has $2.72$ Inception Score (IS) on the CelebA dataset of $128\times128$px resolution~\cite{miyato2018spectral}. 
SNGAN~\cite{miyato2018spectral} comprises $17$ layers, including one fully-connected layer at the start, one convolution layer at the end, and $5$ residual blocks.
Each residual block consists of $3$ convolution layers~\cite{2021mimicry}.

We partition the inversion of the entire SNGAN model into inversions of the first fully-connected layer, the last convolution layer, and each of $5$ residual blocks.
For mathematical consistency with prior GAN literature, we use $\mathcal{G}: \mathbf{z}\rightarrow \mathbf{x}$ to indicate the target generator. $\mathbf{z}$ represents the real latent code that generates $\mathbf{x}=\mathcal{G}(\mathbf{z})$.
We aim to learn $\mathcal{G}^{-1}$ that faithfully reconstructs $\mathbf{z}$ given $\mathbf{x}$. 
To evaluate the quality of inversion, we use the inverted latent code $\mathbf{z}'$ to re-generate an image $\mathbf{x}'=\mathcal{G}(\mathbf{z}')$, which is referred to as the second generation.

The adversary can manipulate the second generation by varying the inverted latent code for 
downstream attacks~\cite{xia2022gan}.

\input{figs/gan_inv/gan_inter}
\noindent  For instance, an adversary may perform linear interpolation between inverted latent codes $\alpha\cdot\mathbf{z'_1} + (1-\alpha)\cdot\mathbf{z'_2}$ with different factor $\alpha\in\{0,0.2,0.4,0.6,0.8,1.0\}$ in~\cref{fig:gan_int} to get manipulated images. In addition, results of~\cref{fig:gan_int} validate that the inverted latent codes fit into the original latent code distribution, akin to GAN inversion literature~\cite{QiLei2019,xia2022gan}.
\input{figs/gan_inv/gan_inv_figs}

% \vspace{-2mm}
\subsection{Trade-off between Utility and Defense}
We next demonstrate that it is quite challenging to defend against the proposed inversion attack by perturbing features while preserving the accuracy of target model. Following prior work of split computing~\cite{teerapittayanon2017distributed,verbraeken2020survey,dong2022splitnets}, we use the first $35$ layers of ResNet-50 as the on-device sub-network and perform inversion attack to the features generated by this sub-network.
Three widely used defense strategies are tested as follows.

\noindent\textbf{Noise Injection.} One straightforward attempt to defend inversion attack is to add noise on features before sharing~\cite{pettai2015combining}. We experiment with Gaussian and Laplacian noises which are widely used in defenses of inversion attacks ~\cite{vepakomma2018no,titcombe2021practical} and differential privacy studies~\cite{abadi2016deep,zhu2020private,gawron2022feature}. We refer to the standard deviation of a noise distribution as noise level and increase it gradually to observe the trade-off between utility and defense, which are measured by the accuracy of the target model and $\frac{100}{\text{PSNR}}$, respectively. A greater $\frac{100}{\text{PSNR}}$ value indicates increased resistance to inversion attack. we observe that the
trade-off between utility and defense mainly depends on the magnitude of noise levels and less related to the noise types. In addition, both types of noises will introduce a non-negligible accuracy drop for the target model. For instance, after adding Laplacian noise with the scale of $0.2$, the recovered images are still clearly recognizable. However, the accuracy of target model drops for about $20\%$ because of the noise. 

\noindent\textbf{Feature Dropout.} We also experiment feature dropout defense by randomly setting elements in features to zero~\cite{he2020attacking}.
The results of varying the sparsity level of features are shown in~\cref{fig:defense}. We observe that all defense methods will degrade the target model's accuracy and impede the inversion attack simultaneously. Although feature dropout can better preserve target model's accuracy, it also has less defense effect. 
These findings validate the non-trivial trade-off between utility and defense, and motivates future research on advanced defense. 
\input{figs/diff_methods/diff_methods}
\input{tabs/diff_methods}
\input{figs/defense/defense_fig}

% \vspace{-2mm}
\section{Conclusion}
This work presents Divide-and-Conquer Inversion (DCI) that scales model inversion attack to large-scale datasets (\eg, ImageNet and CelebA), deep architectures (\eg, RepVGG and ResNet), and different model types (\eg, discriminative and generative). 
Our experimental results show that it is still possible to recover well recognizable images given features after up 42 convolutional layers on ImageNet while existing attacks all fail.
We hope the proposed method can serve as a tool
for empirically measuring the amount of data leakage and facilitate the future study on privacy of split computing.

\clearpage
\bibliography{egbib}
\clearpage
\input{appendix}

\end{document}

%% file: figs/sc/sc_fig.tex
\begin{wrapfigure}{r}{0.5\textwidth}
\vspace{-5mm}
\begin{minipage}{0.5\textwidth}
    \centering
    \includegraphics[width=\linewidth]{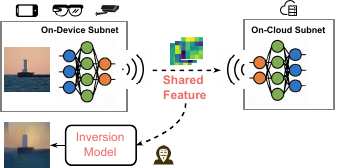}
    \caption{Illustration of model inversion attacks to a DNN under split computing. An adversary retrieves shared feature activation to recover the raw input with an inversion model.}
    \label{fig:teaser}
    \vspace{-3mm}
% \end{figure}
\end{minipage}
\end{wrapfigure}

%% file: tabs/related_work.tex
\begin{table}[t]
\centering
% \vspace{-3mm}
% \vspace{-3mm}
\resizebox{\textwidth}{!}{
\begin{tabular}{lcccccccc}
\toprule
 &\makecell{{PSiM}\\\citep{Dosovitskiy2016,RN58}}& \makecell{IHT\\\citep{Gilbert2017}} & \makecell{INNs\\\citep{behrmann2019invertible_res,behrmann2021understanding}} & \makecell{PlugPlay\\\citep{Nguyenplugplay2017,struppek2022plug,wang2021variational}} & \makecell{DI\\\citep{yin2020dreaming,li2020mixmix}} & \makecell{rMSE\\\citep{he2019model}} & \makecell{GMI\\\citep{zhang2020secret}} & \textbf{Ours} \\
\midrule
No requirement of \textbf{real data} & \ding{55} & &  & \ding{55} &   & &&\checkmark  \\
No requirement of \textbf{pretrained GAN}  & & &  &  \ding{55} &  & &&\checkmark \\
No req. of \textbf{specific arch.} or \textbf{weights dist.}  &  & \ding{55} & \ding{55} & & & &&\checkmark \\
Generalize to both \textbf{disc.} and \textbf{gen.} models & \ding{55} & \ding{55} & & \ding{55} & \ding{55} & & \ding{55} &\checkmark \\
No \textbf{adversarial training} & \ding{55} &&& \ding{55} &&& \ding{55} & \checkmark\\
Fast (\eg, \textbf{single-pass}) inversion &  &  & \ding{55} & & \ding{55} & \ding{55} & & \checkmark \\
\textbf{Sample-specific} inversion & & & & \ding{55} & \ding{55} &  & \ding{55} & \checkmark\\
\bottomrule
\end{tabular}
}
\caption{Comparison between our approach and prior work.
}
\label{tab:method_comparison}
\end{table}

%% file: figs/moco_inv/only_resnet50.tex
\begin{figure}[!t]
\centering
% \vspace{-8mm}
%%% please modify here

\begingroup
\begin{tabular}{c}
% \includegraphics[width=0.45\linewidth,clip,trim=5px 0 0 4px]{figs/real_inv/inv/merged_inv.png}
% \\
\includegraphics[width=0.8\linewidth]{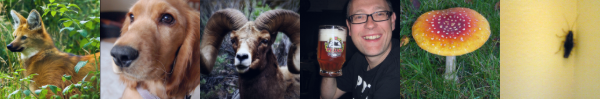} \\
\end{tabular}
\endgroup
% \vspace{1mm}

\small{\ \ Real images $\mathbf{x}$ from the ImageNet validation set.} \\

\begingroup
\begin{tabular}{c}
\includegraphics[width=0.8\linewidth]{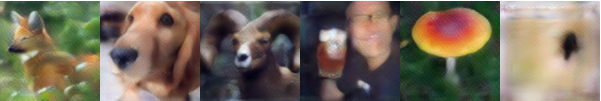} \\
\end{tabular}
\endgroup

\small{Inverted images from ResNet-50's (supervised~\cite{He2015}) features after $42$ conv. layers.
} \\

\begingroup
\begin{tabular}{c}
\includegraphics[width=0.8\linewidth]{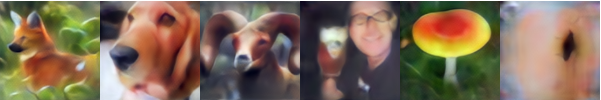} \\
\end{tabular}
\endgroup

\small{Inverted images from ResNet-50's (MoCoV2~\citep{chen2020mocov2}) features after $42$ conv. layers.
} \\
\vspace{2mm}
\caption{Inversion attacks to ResNet-50 (sup. and self-sup.) without access to any real data.
}
\vspace{-4mm}
\label{fig:moco_main}
\end{figure}

%% file: figs/repvgg_main/repvgg_main.tex
\begin{wrapfigure}{r}{0.65\textwidth}
\vspace{-5mm}
\begin{minipage}{0.65\textwidth}

\begin{figure}[H]
\centering
% \vspace{-8mm}
%%% please modify here

\begingroup
\begin{tabular}{c}
% \includegraphics[width=0.45\linewidth,clip,trim=5px 0 0 4px]{figs/real_inv/inv/merged_inv.png}
% \\
\includegraphics[width=0.95\linewidth]{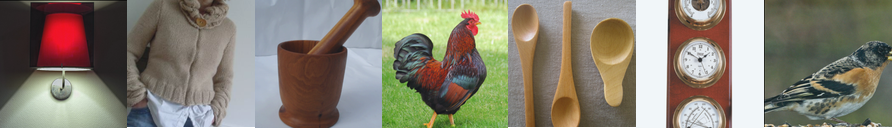} \\
\end{tabular}
\endgroup
% \vspace{1mm}

\small{Real images $\mathbf{x}$ of $224\times224$ px. from the ImageNet validation set.} \\

\begingroup
\begin{tabular}{c}
\includegraphics[width=0.95\linewidth]{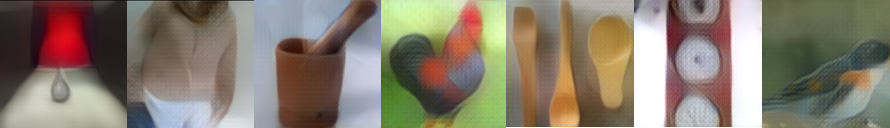} \\
\end{tabular}
\endgroup
\small{Recovered images from features after 21 blocks \scalebox{0.8}{$\mathcal{F}_{1:21}^{-1}(\mathcal{F}_{1:21}(\mathbf{x})\big)$}}. \\
\caption{Inversion Attack to a pre-trained RepVGG. 
}
\vspace{-4mm}
\label{fig:repvgg_main}
\end{figure}

\end{minipage}
\end{wrapfigure}

%% file: figs/layerwise_repvgg/layerwise_repvgg_fig.tex
\begin{wrapfigure}{r}{0.65\textwidth}
\vspace{-3mm}
\begin{minipage}{0.65\textwidth}
% \begin{figure}
    \centering
    \includegraphics[width=\linewidth]{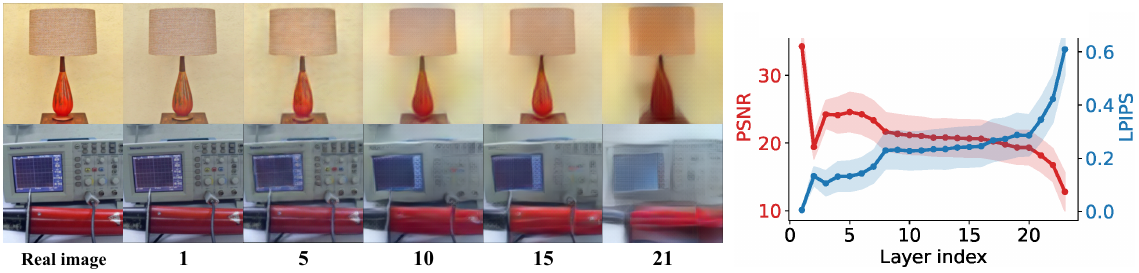}
    \caption{Inversion from features at increased depths. (\textit{Left}) Number indicates layer depth. (\textit{Right}) PSNR and LPIPS at various layer depths. Shaded region is the standard deviation across samples from ImageNet validation set.  
    }
    \label{fig:diff_layer_repvgg}
    \vspace{-4mm}
% \end{figure}
\end{minipage}
\end{wrapfigure}

%% file: figs/resnet18_inv/resnet18_inv.tex
\begin{wrapfigure}{r}{0.65\textwidth}
\vspace{-6mm}
\begin{minipage}{0.65\textwidth}

\begin{figure}[H]
\centering
% \vspace{-8mm}
%%% please modify here

\begingroup
\begin{tabular}{c}
% \includegraphics[width=0.45\linewidth,clip,trim=5px 0 0 4px]{figs/real_inv/inv/merged_inv.png}
% \\
\includegraphics[width=0.97\linewidth]{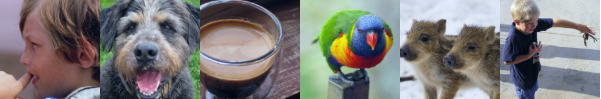} \\
\end{tabular}
\endgroup
% \vspace{1mm}

\small{\ \ Real images from the ImageNet validation set.} \\

\begingroup
\begin{tabular}{c}
\includegraphics[width=0.97\linewidth]{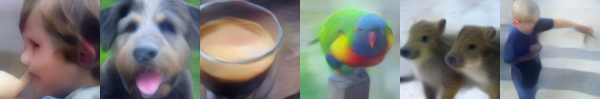} \\
\end{tabular}
\endgroup

\small{Inverted images after $15$ conv. layers \scalebox{0.8}{$\mathcal{F}_{1:15}^{-1}(\mathcal{F}_{1:15}(\mathbf{x})\big)$}}. \\
\caption{\ \ Inversion attack to a pre-trained ResNet-18. 
% Note that recovered images have contextually correct backgrounds, in realistic scenarios, of close proxy to real samples.
}
\vspace{-4mm}
\label{fig:resnet18_main}
\end{figure}

% \vspace{-4mm}
% \input{figs/moco_inv/moco_inv}
% \vspace{-5mm}
\end{minipage}
\end{wrapfigure}

%% file: figs/gan_inv/gan_inter.tex
\begin{wrapfigure}{r}{0.65\textwidth}
\vspace{-12mm}
\begin{minipage}{0.65\textwidth}

\begin{figure}[H]
    \centering
    \includegraphics[width=0.95\linewidth]{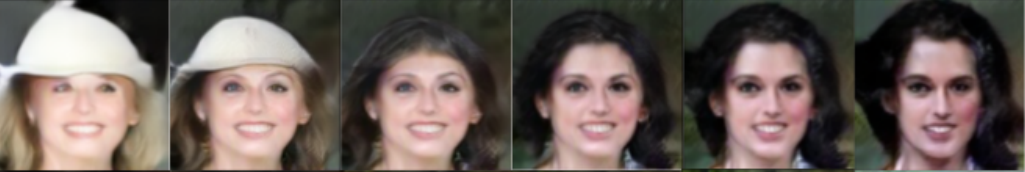}
    \caption{Images from interpolation of inverted latent codes \scalebox{0.8}{$\mathcal{G}(\alpha\cdot\mathbf{z}'_1+(1-\alpha)\cdot\mathbf{z}'_2)$}.}
    \label{fig:gan_int}
    % \vspace{-2mm}
\end{figure}

\end{minipage}
\vspace{-4mm}
\end{wrapfigure}

%% file: figs/gan_inv/gan_inv_figs.tex
\begin{figure}[!t]
\centering
% \vspace{-8mm}
%%% please modify here

\begingroup
\begin{tabular}{c}
% \includegraphics[width=0.45\linewidth,clip,trim=5px 0 0 4px]{figs/real_inv/inv/merged_inv.png}
% \\
\includegraphics[width=0.9\linewidth]{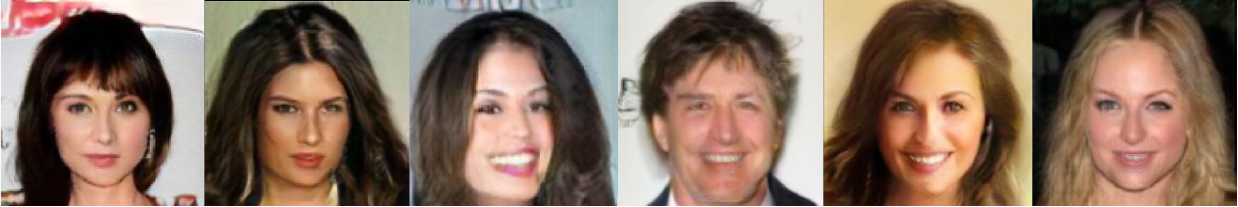} \\
% \vspace{-1mm}
\end{tabular}
\endgroup

\small{SNGAN generated samples \scalebox{0.8}{$\mathbf{x}=\mathcal{G}(\mathbf{z})$}.} \\

\begingroup
\begin{tabular}{c}
\includegraphics[width=0.9\linewidth]{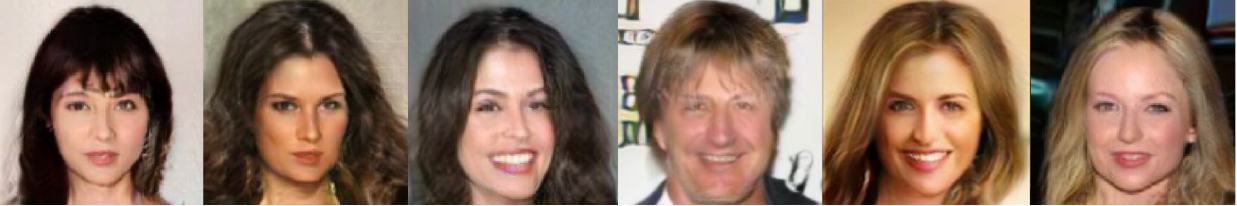} \\
\end{tabular}
\endgroup

\small{Inverted images \scalebox{0.8}{$\mathbf{x}'=\mathcal{G}(\mathcal{G}^{-1}(\mathcal{G}(\mathbf{z})))$}}. \\
\caption{Inversion attack to SNGAN.}
\vspace{-6mm}
\label{fig:inv_sngan}
\end{figure}

%% file: figs/diff_methods/diff_methods.tex
\begin{figure}[t]
\centering
\resizebox{1\linewidth}{!}{
\begingroup
\renewcommand*{\arraystretch}{0.3}
\begin{tabular}{ccccccc}
\includegraphics[width=0.3\linewidth,clip,trim=5px 0 0 4px]{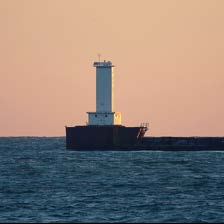} &
\includegraphics[width=0.3\linewidth,clip,trim=5px 0 0 4px]{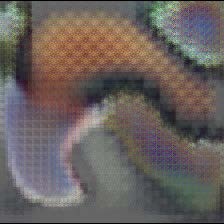} &
\includegraphics[width=0.3\linewidth,clip,trim=5px 0 0 4px]{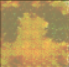} &
\includegraphics[width=0.3\linewidth,clip,trim=5px 0 0 4px]{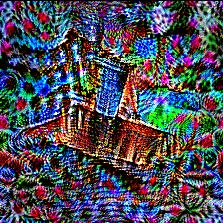} &
\includegraphics[width=0.3\linewidth,clip,trim=5px 0 0 4px]{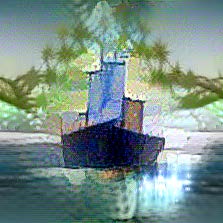} &
\includegraphics[width=0.3\linewidth,clip,trim=5px 0 0 4px]{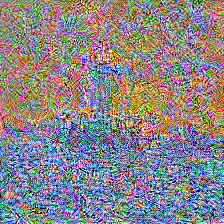} &
\includegraphics[width=0.3\linewidth,clip,trim=5px 0 0 4px]{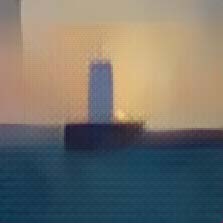} 
\\[2mm]
 \LARGE{Real} & \LARGE{PSiM~\cite{RN58}} & \LARGE{GMI~\cite{zhang2020secret}} & \LARGE{DDream~\cite{mordvintsev2015deepdream}}  & \LARGE{DI~\cite{yin2020dreaming}} & \LARGE{rMSE~\cite{he2019model}} & \LARGE{\textbf{Ours}}
\\
\end{tabular}
\endgroup
}
% \vspace{1mm}
\caption{Qualitative comparison on RepVGG-A0 inversion under data-free setting. 
% Our method recovers the input faithfully.
}
\vspace{-2mm}
\label{fig:sota_comp}
\end{figure}

%% file: tabs/diff_methods.tex
\begin{table}[t]
\centering

% \vspace{-3mm}
\resizebox{\linewidth}{!}{
    \begin{tabular}{ccccccc}
    \toprule
    Method  & PSiM~\cite{RN58} & GMI~\cite{zhang2020secret} & rMSE~\cite{he2019model}  & DDream~\cite{mordvintsev2015deepdream}  & DI~\cite{yin2020dreaming} & \textbf{Ours}\\
    \midrule
    MSE $\downarrow$ & $0.092$ & $0.104$ & $0.082$ & $0.186$ & 0.080 & $\mathbf{0.013}$\\
    LPIPS $\downarrow$ & $0.725$ & $0.751$ & $1.082$ & $0.904$ & $0.602$ & $\mathbf{0.443}$\\
    % FFT-2D $\downarrow$ & 0.271 & 0.131 & 0.174 & 0.153 & 0.231 & $\mathbf{0.009}$\\
    PSNR $\uparrow$ & $9.080$ & $9.395$ & $10.90$ & $9.534$ & $10.93$ & $\mathbf{18.83}$ \\    
    Inference$^\dagger$ (s) $\downarrow$ &$2.4\times 10^{-1}$ & $7.2\times 10^{-2}$ & $2.1\times 10^2$ & $5.5\times 10^2$ & $1.9\times10^3$ & $\bm{1.5\times10^{-2}}$  \\
    \bottomrule
    \multicolumn{4}{l}{\footnotesize{$^\dagger$ Inference time measured on NVIDIA V100 GPU for one inversion of batch size $4$.}}
    \end{tabular}
    }
    % \vspace{0.5mm}
    \caption{Quantitative comparison to prior methods under the data-free setting.}
    \label{tab:comp_number}
    % \vspace{-4mm}
\end{table}

%% file: figs/defense/defense_fig.tex
\begin{figure}[!t]
\vspace{-3mm}
    \centering
    \includegraphics[width=0.95\linewidth]{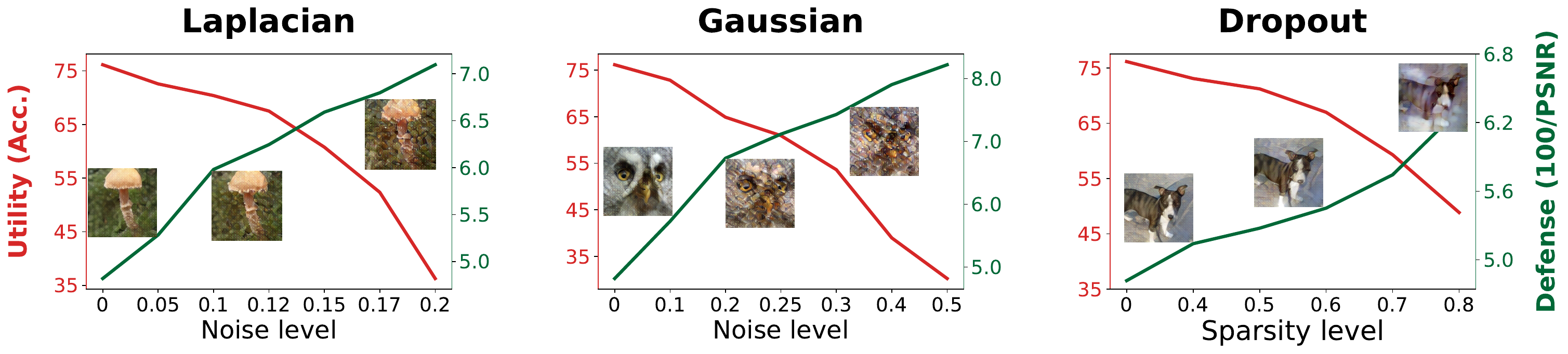}
    \caption{The utility and defense trade-off of defense strategies. Red (green) lines are the accuracy (defense effect $\frac{100}{\text{PSNR}}$) when different noise/sparsity levels are applied to features.}
    \label{fig:defense}
    \vspace{-4mm}
\end{figure}

%% file: appendix.tex
\appendix

\section{Additional Related Work}
\fakeparagraph{Autoencoder.} Model inversion is also related to autoencoders~\cite{vincent2010stacked,hinton2011transforming} given a similar functionality of target and inversion models, to an encoder-decoder pair.
However, the encoder and decoder in an autoencoder are trained jointly with a vast amount of data in an end-to-end manner. In our case, only an individually pre-trained target model is accessible. 
Aligning with the previous finding of layer-wise training of autoencoder (\textit{e.g.}, deep belief network~\cite{hinton2006fast,bengio2007greedy}), we also find layer-wise training beneficial for inversion model training as detailed in~\cref{sec:dci}.

% \vspace{2mm}
\fakeparagraph{Invertible Neural Networks.}
Invertible neural networks (INNs)
\cite{behrmann2021understanding,dinh2016density,kingma2018glow} are a family of neural networks which is theoretically invertible due to their special architectures and weights. 
For instance, i-RevNet~\cite{jacobsen2018revnet} uses carefully planned convolution, reshuffling, and partitioning procedures to ensure inversion. i-ResNet~\cite{behrmann2019invertible_res} constrains weights with Lipschitz condition. 
In conclusion, INNs are inapplicable to standard model architectures since they depend on uncommon architectures and regularizations.

\begin{table}
    \centering
    \caption{Key differences between Gradient Inversion and Feature Inversion (Ours)}
    \label{tab:key_difference}
    % \vspace{-8pt}
    \resizebox{\columnwidth}{!}{%
    \begin{tabular}{|C{0.15\columnwidth}|L{0.5\columnwidth}|L{0.7\columnwidth}|}
         \hline
          & \multicolumn{1}{c|}{\textbf{Gradient Inversion}~\cite{yin2021see,zhu2019deep,li2022auditing,geiping2020inverting,hatamizadeh2022gradvit}} & \multicolumn{1}{c|}{\textbf{Feature Inversion (Ours)}} \\\hline
         \bf Scenario & Federated learning (\textbf{\textit{training}}) & Split computing  \textbf{\textit{(inference)}} \\\hline
          \rotatebox[origin=c]{0}{\bf Input} & Weights + Gradients & Weights + Features\\\hline
          \rotatebox[origin=c]{0}{\bf Goal} & Exact image recovery from \textit{\textbf{gradients}} & Exact image recovery from \textit{\textbf{features}} \\\hline
          
          \bf Insight & Gradients encode private information via inversion & Features encode private information via \textit{even faster} inversion \\\hline
    \end{tabular}
    }
    % \vspace{-7mm}
\end{table}

\section{Implementation Details}
\label{sec:implement_details}
\subsection{Training Strategy}
\label{sec:training strategy}
DCI divides the computation of a feed-forward neural network into multiple parts and inverts the computational flow progressively. One straightforward strategy is to sequentially optimize each individual inversion module $\mathcal{F}_k^{-1}$, beginning with the first layer $\mathcal{F}_1^{-1}$. 
We observe that as we progress deeper into the model, the inversion error will increase. To address this accumulation issue, we use a more potent training strategy.
After optimizing a particular inversion module $\mathcal{F}_k^{-1}$, we take into account all previous inversion modules and fine-tune all layers up to $k$, \textit{i.e.,}  $\mathcal{F}_1^{-1}\circ\dots\circ \mathcal{F}_{k-1}^{-1}\circ \mathcal{F}_k^{-1}$, with the same loss $\mathcal{L}_k$ to reduce the accumulated inversion error. When $k=1$, this fine-tuning is skipped because there is no inversion model prior to $\mathcal{F}_1^{-1}$.

\subsection{Inversion Models}
\fakeparagraph{ResNet-18.}
The first four sequential layers in ResNet models are convolution, BN, ReLU, and max pooling layers, and we invert the sequence of the first four layers as a whole block. Each inversion module mimics the architecture of its corresponding target counterpart but with reversed input and output dimensions. For a \texttt{BasicBlock} (left) in ResNet-18, its inversion module (right) is shown as below in Fig.~\ref{fig:app_schematic}.
\begin{figure}
    \centering
    \quad\quad\includegraphics[width=0.8\linewidth]{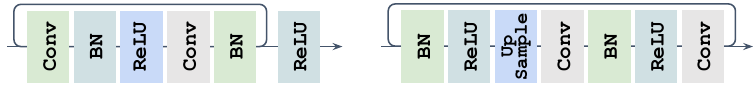}
    \caption{(Left) A \texttt{BasicBlock} in ResNet-18 (Right) Corresponding inversion module for the \texttt{BasicaBlock}.}
    \label{fig:app_schematic}
    % \vspace{-4mm}
\end{figure}

\vspace{2mm}
\fakeparagraph{ResNet-50.}
We break the overall ResNet-$50$ architecture into five sub-networks and invert one sub-network each time. Similar to ResNet-$18$, the first sub-network is the initial block. The other four sub-networks consist of $3,4,6,3$ repeats of \texttt{Bottleneck} respectively~\cite{He2015}. 

\section{Additional Inversion Results}
\label{sec:addition_inversion}
\fakeparagraph{Lossy Final Fully-Connected Layer.}
In the main manuscript, we show that the recovered images from RepVGG feature embeddings after $21$ convolution layers preserve original semantic and visual attributes. However, we find that information in the feature embeddings decays rapidly through the last two layers in RepVGG. This may indicate that class-invariant information is quickly filtered out towards the end of the model while the initial stages focus on feature extraction, which is aligned with prior observations in transfer learning~\cite{NIPS2014_375c7134,li2020rifle,long2015learning,dollar2018rethinking,zoph2020rethinking} and self-supervised learning~\cite{he2020momentum,grill2020bootstrap,chen2020big}. 
We visualize the above findings in~\cref{fig:rep_last_two_layer}. One can still recognize the class of inverted images after the penultimate (\textit{i.e.}, the $22$-th convolution) layer. However, if we invert features after the final (\textit{i.e.}, the fully-connected) layer, only the predominant color is recognizable. 
\input{figs/rep_appendix/rep_appendix}

\section{Ablation Studies}
\label{sec:ablation}
\fakeparagraph{Effectiveness of Synthetic Data.}
\input{figs/abla_data/data_fig}
We study how well synthetic data may be used to optimize our inversion modules.
For a sanity check, we re-perform the experiment in~\cref{fig:repvgg_main} using real data from ImageNet training set rather than the synthetic data. We also vary the number of synthetic data used for optimization. In~\cref{fig:abla_data}, we show that using more (real or synthetic) data samples improves the inversion quality. For instance, in the first column of~\cref{fig:abla_data}, the inverted images can reveal the parrot's tiny eyes when 10K (real or synthetic) images are used for the optimization of inversion modules.  
However, we also discover that when there are more than 10K (real or synthetic) samples, the improvement of inversion quality is limited. In addition, we demonstrate that \method achieves a comparable inversion quality when using synthetic data as opposed to using real data. 
However, due to their sensitivity to data distribution change, as illustrated in~\cref{fig:sota_comp}, prior generative inversion approaches~\cite{Nguyenplugplay2017,zhang2020secret} cannot efficiently use synthetic data.
\input{figs/abla_more_res/abla_more_fig}

%% file: figs/rep_appendix/rep_appendix.tex
\begin{figure}[h]
\centering

\resizebox{1\linewidth}{!}{
\begingroup
\renewcommand*{\arraystretch}{0.3}
\begin{tabular}{ccc}
\includegraphics[width=0.3\linewidth,clip,trim=5px 0 0 4px]{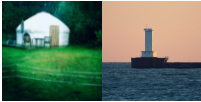} &
\includegraphics[width=0.3\linewidth,clip,trim=5px 0 0 4px]{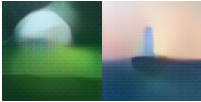} &
\includegraphics[width=0.3\linewidth,clip,trim=5px 0 0 4px]{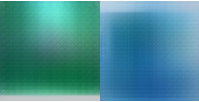} 
\\
 (a) Original images & \makecell{(b) Inversion from the \\ last convolution layer} & \makecell{(c) Inversion from the \\ final full-connected layer}
\\
\end{tabular}
\endgroup
}

\caption{Results of inversion from (b) the penultimate layer~(\textit{i.e.}, the last convolution layer), and (c) the final layer~(\textit{i.e.}, the final full-connected layer)}
\label{fig:rep_last_two_layer}
\end{figure}

%% file: figs/abla_data/data_fig.tex
\begin{figure}[!t]
\centering
% \vspace{-8mm}
%%% please modify here

\begingroup
\begin{tabular}{c}
% \includegraphics[width=0.45\linewidth,clip,trim=5px 0 0 4px]{figs/real_inv/inv/merged_inv.png}
% \\
\includegraphics[width=0.85\linewidth]{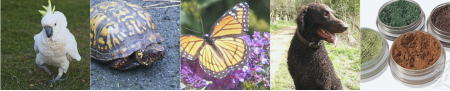} \\
\end{tabular}
\endgroup
% \vspace{1mm}

\small{\ \ Real images $\mathbf{x}$ from the ImageNet validation set.} \\

\vspace{4mm}
\begingroup
\begin{tabular}{c}
\includegraphics[width=0.85\linewidth]{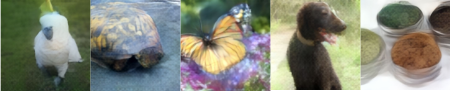} \\
\end{tabular}
\endgroup

\small{Inverted images by inversion model trained on 1K real data.\quad\quad\quad (LPIPS $\downarrow$\ =\ 0.462)} \\

\vspace{4mm}
\begingroup
\begin{tabular}{c}
\includegraphics[width=0.85\linewidth]{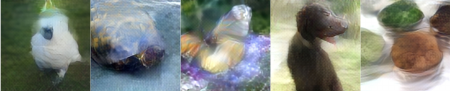} \\
\end{tabular}
\endgroup

\small{Inverted images by inversion model trained on 1K synthetic data.\ \ \ (LPIPS $\downarrow$\ =\ 0.481)} \\

\vspace{4mm}
\begingroup
\begin{tabular}{c}
\includegraphics[width=0.85\linewidth]{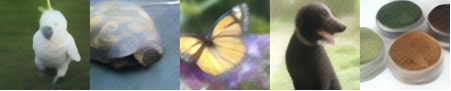} \\
\end{tabular}
\endgroup

\small{Inverted images by inversion model trained on 10K real data.\quad\quad\ \ (LPIPS $\downarrow$\ =\ 0.430)} \\

\vspace{4mm}
\begingroup
\begin{tabular}{c}
\includegraphics[width=0.85\linewidth]{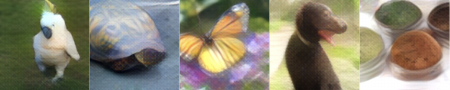} \\
\end{tabular}
\endgroup

\small{Inverted images by inversion model trained on 10K synthetic data. (LPIPS $\downarrow$\ =\ 0.443)} \\
\vspace{4mm}
\caption{Comparison of different data settings. Our method achieves a comparable inversion quality using synthetic data compared to using real data.  
}
% \vspace{-4mm}
\label{fig:abla_data}
\end{figure}

%% file: figs/abla_more_res/abla_more_fig.tex
\begin{figure}[!t]
\centering
% \vspace{-8mm}
%%% please modify here

\begingroup
\begin{tabular}{c}
% \includegraphics[width=0.45\linewidth,clip,trim=5px 0 0 4px]{figs/real_inv/inv/merged_inv.png}
% \\
\includegraphics[width=0.7\linewidth]{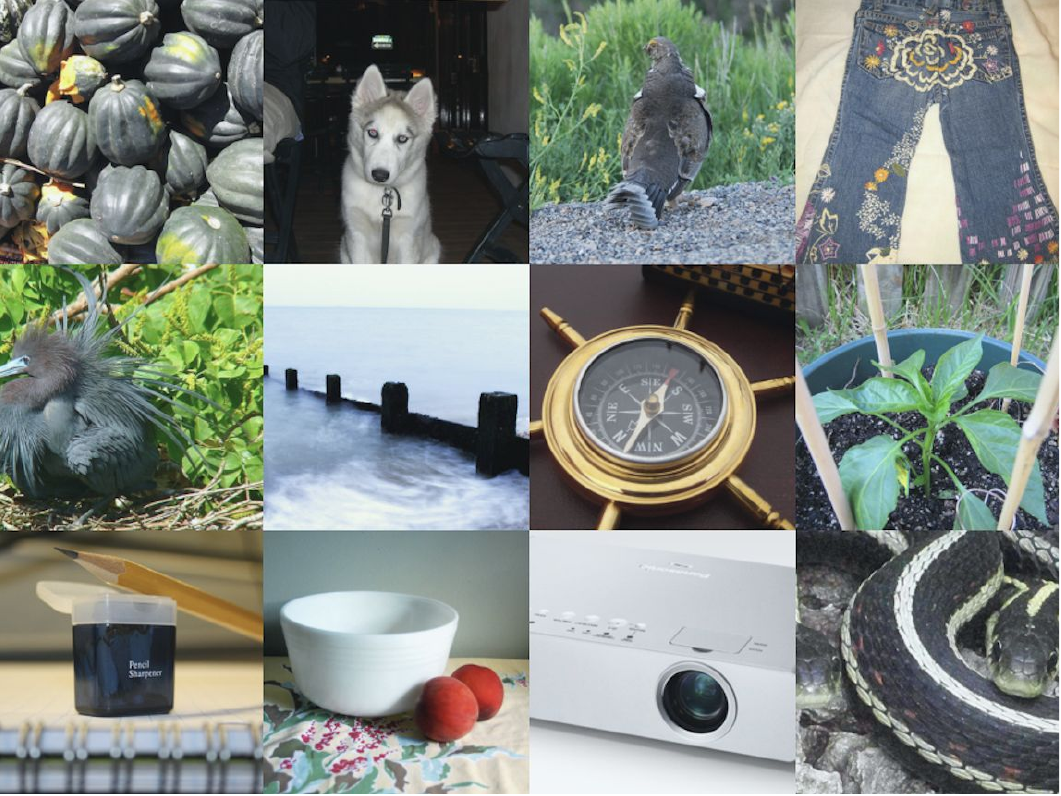} \\
\end{tabular}
\endgroup
% \vspace{1mm}

\small{\ \ Real images $\mathbf{x}$ from the ImageNet validation set.} \\

\vspace{12mm}

\begingroup
\begin{tabular}{c}
\includegraphics[width=0.7\linewidth]{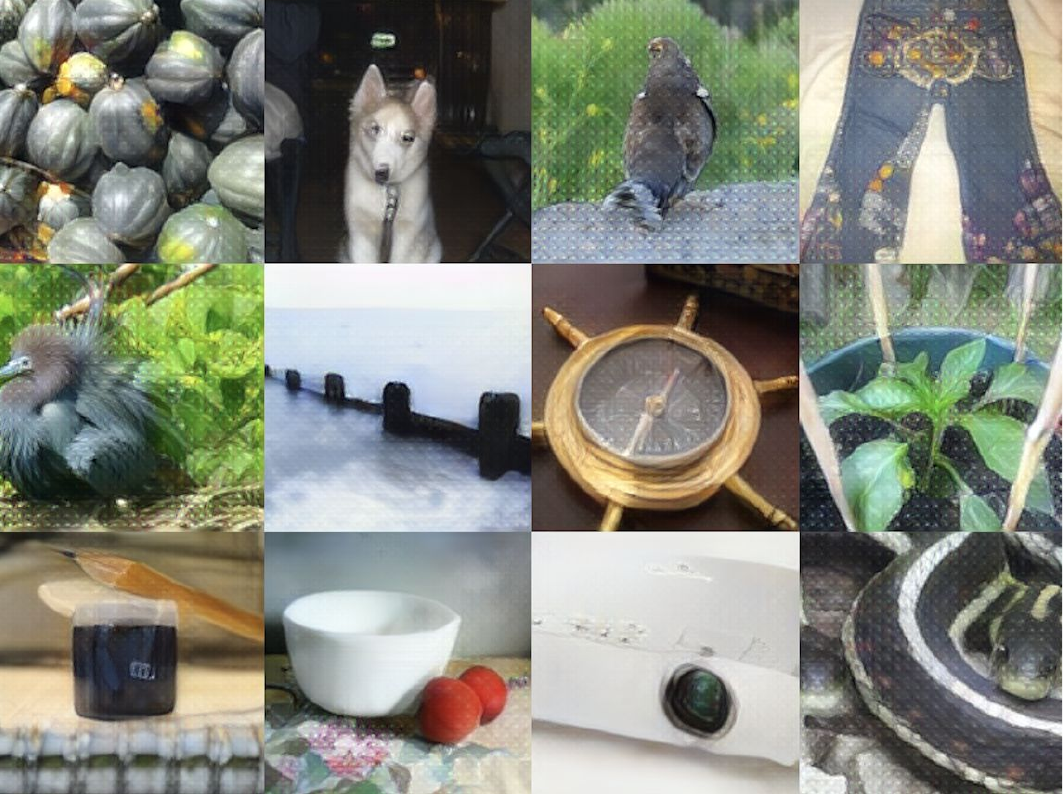} \\
\end{tabular}
\endgroup

\small{Inverted images from ResNet-50’s (supervised) features after 35 conv. layers.} \\
\vspace{4mm}
\caption{More inversion results on ResNet-50 without any real data. Our method generalizes well on different types of images. 
}
\vspace{-4mm}
\label{fig:abla_more_res}
\end{figure}

\begin{figure}[!t]
\centering
% \vspace{-8mm}
%%% please modify here

\begingroup
\begin{tabular}{c}
% \includegraphics[width=0.45\linewidth,clip,trim=5px 0 0 4px]{figs/real_inv/inv/merged_inv.png}
% \\
\includegraphics[width=0.7\linewidth]{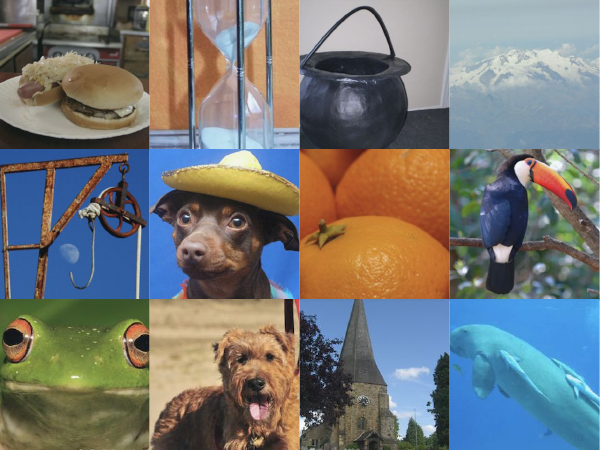} \\
\end{tabular}
\endgroup
% \vspace{1mm}

\small{\ \ Real images $\mathbf{x}$ from the ImageNet validation set.} \\

\vspace{12mm}

\begingroup
\begin{tabular}{c}
\includegraphics[width=0.7\linewidth]{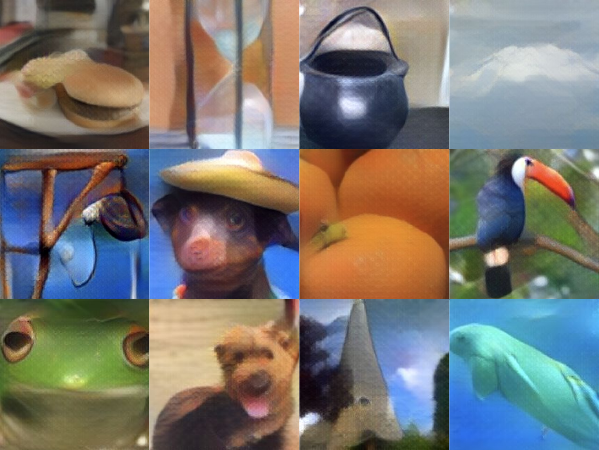} \\
\end{tabular}
\endgroup

\small{Inverted images from RepVGG-A0 features after 21 conv. layers.} \\
\vspace{4mm}
\caption{More inversion results on RepVGG-A0 without any real data. Our method generalizes well on different types of images. 
}
\vspace{-4mm}
\label{fig:abla_more_res}
\end{figure}